\documentclass[runningheads]{llncs}
\usepackage{graphicx}%
\usepackage{multirow}%
\usepackage{amsmath,amssymb,amsfonts}%
\usepackage{mathrsfs}%
\usepackage[title]{appendix}%
\usepackage{xcolor}%
\usepackage{textcomp}%
\usepackage{manyfoot}%
\usepackage{booktabs}%
\usepackage{algorithm}%
\usepackage{algorithmicx}%
\usepackage{algpseudocode}%
\usepackage{listings}%
\usepackage{changepage}
\usepackage[T1]{fontenc}
\usepackage{comment}
\usepackage{cite}

\begin{document}
\title{EMG-Based Hand Gesture Recognition through Diverse Domain Feature Enhancement and Machine Learning-Based Approach}
\author{Abu Saleh Musa Miah\inst{1}\orcidID{0000-0002-1238-0464} \and
Najmul Hassan\inst{1}\orcidID{0009-0000-6499-1825} \and
Md. Maniruzzaman \inst{1}\orcidID{0000-0001-6151-8071}\and
Nobuyoshi Asai \inst{1}\and
Jungpil Shin \inst{1}\inst{*}\orcidID{0000-0002-7476-2468}
}

\institute{School of Computer Science and Engineering,
The University of Aizu, Aizuwakamatsu, Fukushima, Japan.\\ \email{musa@u-aizu.ac.jp}, *\email{jpshin@u-aizu.ac.jp}}

%\email{jpshin@u-aizu.ac.jp}
%
\maketitle              % typeset the header of the contribution
\begin{abstract}
Surface electromyography (EMG) serves as a pivotal tool in hand gesture recognition and human-computer interaction, offering a non-invasive means of signal acquisition. This study presents a novel methodology for classifying hand gestures using EMG signals. To address the challenges associated with feature extraction where, we explored 23 distinct morphological, time domain and frequency domain feature extraction techniques. However, the substantial size of the features may increase the computational complexity issues that can hinder machine learning algorithm performance. To mitigate this, we employ an efficient feature selection approach, specifically an extra tree classifier. The selected potential feature fed into the various machine learning-based classification algorithms where our model achieved 97.43\% accuracy with the KNN algorithm and selected feature. By leveraging a comprehensive feature extraction and selection strategy, our methodology enhances the accuracy and usability of EMG-based hand gesture recognition systems. The higher performance accuracy proves the effectiveness of the proposed model over the existing system. 
\end{abstract}

\section{Introduction}
%Definition:
The electrical activity of the skeletal muscles generates electromyography (EMG) signals, which can be captured in two ways: invasive and noninvasive methods \cite{ja2020identification_main,miah2024effective}. These kinds of signals is one of the most crucial ways to implement human-machine communication based on hand gestures or human activity information \cite{phinyomark2018feature,42miah2020motor}. %Importance of EMG-based Hand Gesture Recognition:
EMG-based hand gesture recognition is crucial for various applications, including human-computer interaction, prosthetics, rehabilitation, and biomedical research. It enables individuals to control devices and interfaces using natural hand movements, improving accessibility and quality of life \cite{khushaba2012toward}.
%Applications:
EMG signals are utilized in systems for movement classification \cite{ji2017stationary,ghofrani2018cross}, recognition of diagnosis of neuromuscular diseases \cite{kocer2017classifying} and motor unit action potential (MUAP) \cite{benazzouz2019emg}.
%Example:
For instance, EMG signals are used to recognize finger movements for controlling prosthetic devices or computer interfaces, facilitating intuitive interaction for individuals with limb disabilities \cite{phukan2019finger}.
%Existing Research Work:
Many researchers have been working to develop EMG-based hand gesture recognition using various statistical and mathematical formula-based features.  Among them, Hudgins et al. \ cites {tabares2019deep} extracted 4-time domain features and then fed them into the classification approach that produced good performance accuracy compared to the previous study. However, challenges such as changes in fatigue, arm posture, body movement, and natural variability of the human body may also impact the feature effectiveness \cite{orozco2019systematic,10452793_multi_culture}. 
More recently, several studies have been done and compared with other feature extraction excluded time and frequency-domain features, but simpler time-domain features remain robust in classification tasks \cite{subasi2018automated,subasi2020emg}.
More recently, Rubio et proposed an EMG-based hand gesture recognition method with various machine learning models, and they reported 95.39\% accuracy \cite{ja2020identification_main}. The drawback of their model is that it did not achieve satisfactory performance accuracy due to a lack of feature effectiveness.  To increase performance accuracy, it is important to extract the effective feature using various statistical and mathematical formulas. To overcome the problem, we proposed morphological, time domain and frequency domain feature-based EMG-based hand gesture recognition with effective feature selection and classification modules.

The proposed methodology contributes to the field of hand gesture classification using EMG signals in several key ways:
\begin{itemize}
    \item 
    \textbf{Comprehensive Feature Extraction:} The paper introduces the novel feature extraction approach by employing morphological, time domain, and frequency domain feature extraction techniques. This comprehensive approach addresses the challenge of capturing diverse aspects of the EMG signal, enhancing the richness of information available for classification.
    
    \item 
    \textbf{Effective Feature Selection:} To mitigate the computational complexity and reduce the cost associated with large datasets, the methodology incorporates an Extra tree classifier (ETC) algorithm for feature selection. This step helps identify the most relevant features for classification, improving the efficiency of subsequent analysis.
    
    \item 
    \textbf{Classification Approach:} The methodology applies various machine learning algorithms for classification with the original features and selected features. By exploring different algorithms, the study identifies models that yield high classification accuracy, contributing to the advancement of classification techniques.
    
    \item 
    \textbf{Extensive Evaluation:} The experimental results demonstrate the effectiveness of the proposed approach, with classification accuracy exceeding 97.43\% with the selected features using the KNN approach. This performance surpasses the existing state-of-the-art models in terms of performance accuracy and efficiency.
\end{itemize}

\begin{table}[htbp]
\centering
\caption{Dataset description} \label{tab:dataset}
\begin{tabular}{ll}
\toprule
Label   &      Gesture Name \\
\midrule
0 & Intervals between   gestures \\
1 & Wrist extension \\
2 & Grasped hand \\
3 & Resting hand \\
4 & Wrist flexion \\
5 & Radial deviation \\
6 & Ulnar deviation \\
\bottomrule
\end{tabular}
\end{table}
\section{Related Work}
Many academic and industrial research projects have been done to develop EMG based hand gesture recognition  systems. Some researchers used feature extraction techniques, including autoregressive models \cite{chen2007characterization}, signal entropy measurements \cite{gokgoz2015comparison}, and amplitude of the time and frequency domain-based statistical measurements.  Wavelet transforms are used by some researchers for extracting time-frequency domain features from the EMG-based hand gesture dataset to develop a hand gesture recognition system \cite{singh2015wavelets}. Several studies have been working on the development of HGR systems, which are based on the feature extraction and machine learning (ML) algorithm \cite{24taghizadeh2021finger,33mukhopadhyay2020experimental,35esmaeili2020semi,miah2023dynamic,miah2022bensignnet,miah2023multistage,miahrotation}. 
To improve the performance accuracy, Li et al.\cite{27li2013boosting} proposed an ensemble approach that combined the random forest and hosting method and achieved 80\% accuracy on an sEMG-based hand gesture dataset. More recently, Khushaba et al. employed a hand gesture recognition system with 12 FD and TD features of the EMG signal, and they showed 90\% accuracy using the SVM approach~\cite{20al2013classification}. 
Praveen et al. introduced a feature extraction approach incorporating root mean square and integrated absolute value techniques with KNN and Bayesian pattern classification algorithms. Their method yielded promising results, achieving a recognition accuracy 92\% in sEMG-based gesture recognition tasks \cite{28praveen2018design}. More recently, Azhhiri et al. employed frequency domain (FD) features extracted from different levels of wavelet decomposition, leading to notable enhancements in sEMG-based Hand Gesture Recognition (HGR) when integrated with a multilayer perceptron model \cite{30azhiri2021emg}. Arteaga et al. augmented performance accuracy and efficiency by gathering sEMG signals pertaining to six distinct hand gestures from a cohort of 20 participants and subsequently employing the k-Nearest Neighbors (kNN) algorithm \cite{31arteaga2020emg}. More recently, Rubio et proposed an EMG-based hand gesture recognition method with various machine learning models, and they reported 95.39\% accuracy \cite{ja2020identification_main}. The drawback of their model is that it did not achieve satisfactory performance accuracy due to a lack of feature effectiveness.  It is important to extract the effective feature using various statistical and mathematical formulas to increase performance accuracy. To overcome the problem, we proposed morphological, time domain and frequency domain feature-based EMG-based hand gesture recognition with effective feature selection and classification modules.

\section{Dataset}
We collected a free EMG-based hand gesture dataset, which is available in the UCI library \cite{misc_emg_data_for_gestures_481}. The dataset consisted of six static hand gestures collected from 36 patients and a total of 72 recordings. Two MYO thalamic bracelet devices, which have eight channels, were used to collect the dataset. 
The set of static hand gestures encompassed wrist extension, grasped hand, resting hand, wrist flexion, radial deviation and ulnar deviation which are shown in Table \ref{tab:dataset}. The dataset of each gesture was recorded for 3 and 3 seconds for the duration and interval between gestures, respectively. \cite{ja2020identification_main}.
\begin{figure}[htbp]
\begin{adjustwidth} {-3cm}{0cm}
    \centering
\caption{{Workflow architecture of the proposed methodology}.}\label{fig:proposed_main}
\includegraphics[scale=.30]{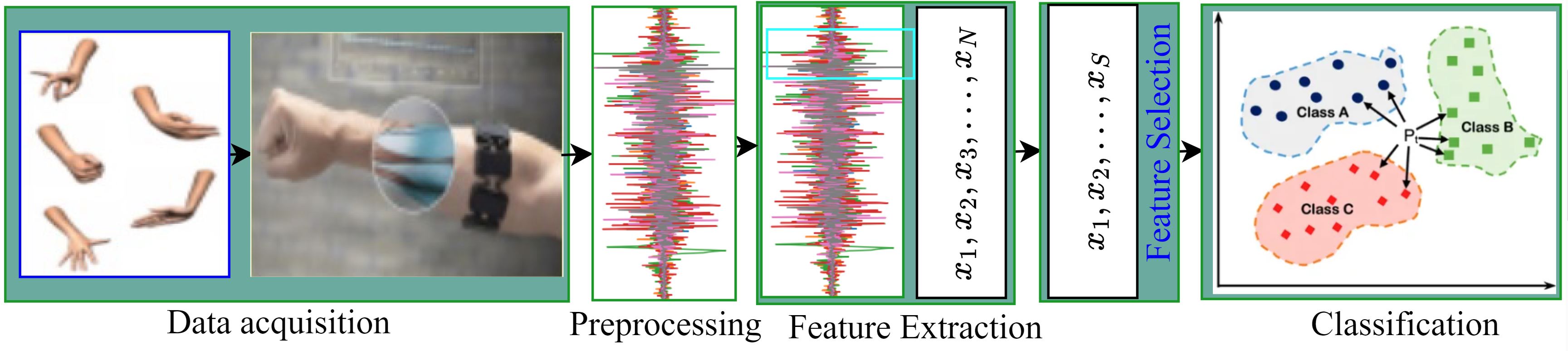}
\end{adjustwidth}
\end{figure}
\section{Proposed Methodology}
This study proposes an innovative methodology for hand gesture classification using surface electromyography (EMG) signals. Figure \ref{fig:proposed_main} demonstrated the proposed methodology where we included several key components: preprocessing, feature extraction, feature selection and classification module. Firstly, we employ 23 different morphological, time-domain and frequency-domain feature extraction techniques to tackle feature extraction challenges from the EMG signals. We extracted these features to capture diverse aspects of the signal's characteristics, thereby enhancing the discriminative power of the feature set. In addition, to reduce the computational complexity, we employed an effective feature selection approach to mitigate these challenges, which leads to increased performance accuracy and model efficiency as well.  By applying a combination of feature extraction, selection, and classification techniques, our approach aims to optimize the overall classification process. We leverage various machine learning algorithms to classify EMG hand gestures here. These algorithms are chosen based on their suitability for handling high-dimensional data and their capacity to generalize well across different subjects. By prioritizing this information, we aim to identify patterns associated with six distinct gestures, irrespective of the individual acting. This generalization among different subjects addresses a significant challenge in EMG-based hand gesture recognition. The proposed methodology is evaluated using EMG recordings from 36 patients performing six hand gestures. 

\subsection{Preprocessing}
During preprocessing, irrelevant EMG signal segments denoted as zero, representing resting intervals between gestures, were eliminated to focus solely on hand movement data. Instead of feature extraction, raw data from the device's eight channels were directly used. This approach aimed to exploit the inherent discriminatory information in raw EMG signals, maintaining data richness and granularity to enhance classification accuracy \cite{ja2020identification_main,miah2024effective}.
\begin{table}[htbp]
\begin{adjustwidth}{-3cm}{0cm}
\caption{Morphological features description} \label{tab:feat_morphological_feature}
\begin{tabular}{lll} 
\toprule
Feature Name & Definition & Formula \\
\midrule
Absolute Amplitude & \begin{tabular}{p{10cm}} % Adjust the width as per your requirement
Denoted as \(x\), is calculated by taking the absolute \\value of the input signal \(x\). 
\end{tabular}& \( x = \lvert x \rvert \)  \\
Minimum Amplitude & \begin{tabular}{p{8cm}} % Adjust the width as per your requirement
Denoted as $x_{\text{min}}$, is the smallest value present in the input signal $x$. 
\end{tabular}& \( x_{\text{min}} = \min(x) \)  \\
Maximum Amplitude & \begin{tabular}{p{8cm}} % Adjust the width as per your requirement
Denoted as $x_{\text{max}}$, is the largest value present in the input signal $x$. 
\end{tabular}& \( x_{\text{max}} = \max(x) \)  \\
Peak to Peak (PTP) & \begin{tabular}{p{8cm}} % Adjust the width as per your requirement
This range, denoted as PTP, is a feature extraction method for EMG signals. It is calculated as the difference between the maximum amplitude ($x_{\text{max}}$) and the minimum amplitude ($x_{\text{min}}$) 
\end{tabular}& \( \text{PTP} = x_{\text{max}} - x_{\text{min}} \)  \\
 \bottomrule
\end{tabular}
\end{adjustwidth}
\end{table}

\subsection{Feature Extraction}
Effective feature extraction is used to highlight the meaningful structure and pattern of the signal, which is one of the most challenging tasks for hand gesture recognition with EMG signals. We extracted morphological features, time domain features, and frequency domain features we described below:\\
\subsubsection{Morphological Feature}
We calculated the morphological features from the EMG signals based on the following formula that is described in Table \ref{tab:feat_morphological_feature}.\\
 \subsubsection{Time Domain Feature}
 We have employed different statistical approaches to obtain the time domain features from the EMG signal. The details description of this feature is described in Table \ref{tab:feat_time_domain}.
 \begin{table}[htbp]
 \begin{adjustwidth}{-3cm}{0cm}
\caption{Time domain features description} \label{tab:feat_time_domain}
\begin{tabular}{lll} 
\toprule
Feature Name & Definition & Formula \\
\midrule
Arithmetic Mean & \begin{tabular}{p{9cm}} % Adjust the width as per your requirement
A common feature of the time series data is the mean. 
\end{tabular}& \( \mu = \frac{1}{N} \sum_{i=1}^{N} x_i \)  \\
Median & \begin{tabular}{p{8cm}} % Adjust the width as per your requirement
Let $n$ be the number of samples and $\mu$ be the mean of the nth numbers.
, where “n” is the number of items in the set and “th” just means the (n)th number.. 
\end{tabular}& \( m = \frac{(n+1)}{2} \cdot \mu \)  \\

Sample Variance (\(svars\)) & \begin{tabular}{p{8cm}} % Adjust the width as per your requirement
the deviation summed up over the samples and then divided by the sample size. 
\end{tabular}& \( \sigma^2 = \frac{1}{(n-1)} \sum_{i=1}^{n} (x_i - \mu)^2 \)  \\

Population Variance (\(svarsp\)) & \begin{tabular}{p{8cm}} % Adjust the width as per your requirement
It is the same as \(\sigma^2\), with the main difference being that the population variance is calculated based on the total population, whereas sample variance is calculated based on the sample size \(n\)
\end{tabular}& \( sigma^2 = \frac{1}{N} \sum_{i=1}^{N} (x_i - \mu)^2 \)  \\
Mean Absolute Deviation & \begin{tabular}{p{8cm}} % Adjust the width as per your requirement
measure of the average absolute difference between each data point and the mean of the dataset. 
\end{tabular}& \( \text{MAD} = \frac{1}{n} \sum_{i=1}^{n} |x_i - \bar{x}| \) \\

Sample Standard Deviation (\(stds\)) & \begin{tabular}{p{8cm}} % Adjust the width as per your requirement
Denoted as \(\sigma\), is calculated as the square root of the sample variance.
\end{tabular}& \( \sigma = \sqrt{\frac{1}{(n-1)} \sum_{i=1}^{n} (x_i - \mu)^2} \) \\

Population Standard Deviation (\(sstdp\)) & \begin{tabular}{p{8cm}} % Adjust the width as per your requirement
Calculated from population variance
\end{tabular}& \( \sigma = \sqrt{\frac{1}{N} \sum_{i=1}^{N} (x_i - \mu)^2} \) \\

Percentile & \begin{tabular}{p{8cm}} % Adjust the width as per your requirement
Let \(p\)-value be 50 and \(n\) be the total number of samples.
\end{tabular}& \( \text{Percentile} = \frac{p}{100 \times (n + 1)} \) \\

Quantile & \begin{tabular}{p{8cm}} % Adjust the width as per your requirement
calculated with the first quartile. It also known as the lower quartile
\end{tabular}& \( Q1 = \left(\frac{(n+1)}{4}\right)\text{th Term} \) \\

Interquartile Range (\(iqs\))& \begin{tabular}{p{9cm}} % Adjust the width as per your requirement
Calculated as the difference between the upper quartile and \\ the lower quartile. Let
\( Q2 = Q3 - Q1 \) \\
\( \text{Q1} = \left(\frac{1}{4}\right)\left[(N + 1)\right]\text{th term} \), \\
\( \text{Q3} = \left(\frac{3}{4}\right)\left[(N + 1)\right]\text{th term} \) \\
\( n = \text{number of data points} \)
\end{tabular}& \( IQR = Q3 - Q1 \) \\
Skewness & \begin{tabular}{p{8cm}} % Adjust the width as per your requirement
Symmetry of the signal measured with this. Let 3rd and 2nd order central moments of the respective EMG signals $m_3$ and $m_2$ then we can write:\\
\(m_3 = E\left[(x_i - \mu)^3\right] = \frac{1}{n} \sum_{i=1}^{n} (x_i - \mu)^3  \)\\
\(m_2 = E\left[(x_i - \mu)^2\right] = \frac{1}{n} \sum_{i=1}^{n} (x_i - \mu)^2 \)\\

\end{tabular}& \( G_{1} = \frac{m_3}{m_2^{3/2}} \) \\

Kurtosis & \begin{tabular}{p{8cm}} % Adjust the width as per your requirement
Tailness of distribution relative to the normal distribution measurement. Let  4th and 2nd order central moments of the respective EMG signals $m_4$ and $m_2$ then we can write:\\
\( m_4 = E\left[(x_i - \mu)^4\right] = \frac{1}{n} \sum_{i=1}^{n} (x_i - \mu)^4  \) \\
\( m_2 = E\left[(x_i - \mu)^2\right] = \frac{1}{n} \sum_{i=1}^{n} (x_i - \mu)^2 \) \\
\end{tabular}& \( G_2 = \frac{m_4}{m_2^2} \) \\
 \bottomrule
\end{tabular}
\end{adjustwidth}
\end{table}\\
\subsubsection{Frequency Domain Feature} We also extracted frequency-domain features, which are shown in Table \ref{tab:feat_frequency_domain_feature}. 

\begin{table}[htbp]
\begin{adjustwidth}{-3cm}{0cm}
\caption{Frequency domain features description} \label{tab:feat_frequency_domain_feature}
\begin{tabular}{lll} 
\toprule
Feature Name & Definition & Formula \\
\midrule
Energy & \begin{tabular}{p{10cm}} % Adjust the width as per your requirement
It is a major frequency domain feature extraction method. 
\end{tabular}& \( \text{En} = \sum_{i=1}^{N} x_i^2 \)  \\
Power & \begin{tabular}{p{8cm}} % Adjust the width as per your requirement
 \end{tabular}& \( P = \frac{1}{N} \sum_{i=1}^{N} x_i^2 \)  \\
Root Mean Square (RMS) & \begin{tabular}{p{8cm}} % Adjust the width as per your requirement
Denoted as $x_{\text{max}}$, is the largest value present in the input signal $x$. 
\end{tabular}& \( \text{RMS} = \sqrt{\frac{1}{N} \sum_{i=1}^{N} x_i^2}\)
\\
Hjort Parameter Activity (HPA) & \begin{tabular}{p{8cm}} % Adjust the width as per your requirement
\end{tabular}& \( \text{HPA} = \frac{1}{N} \sum_{i=1}^{N} (x_i - \mu)^2 \)  \\
 \bottomrule
\end{tabular}
\end{adjustwidth}
\end{table}

\subsection{Feature Selection With Extra Tree Classifier}
After extracting features from EMG signals, we employed the Extra Trees classifier (ETC) for feature selection due to its robustness and efficiency. Extra Trees introduces significant randomization during node splitting, generating fully randomized trees in some cases. This approach enhances diversity and reduces variance, making it well-suited for feature selection tasks. ETC first divide the features into subsets by splitting to ensure a diverse feature combination. This splitting is performed by criteria like Gini impurity or entropy, with the selected feature and threshold yielding the highest impurity reduction per split. Then, it calculates the importance of each individual feature after constructing all trees; feature importance is computed by averaging impurity reduction values across the ensemble. Finally, Importances are typically normalized or converted to percentages for ranking, aiding in identifying the most relevant features \cite{miah2024effective}.

\subsection{Classification}
We implemented various ML models in the study, including Random Forest, Extra Tree Classifier, SVM, Gaussian NB, Gradient Boosting, HGboosting, Decision Tree and KNN \cite{rubio2020identification,miah2022movie_miah,miah2022natural_EEG,kabir2024exploring_miah,miah2021event_EEG}. \\
\textbf{K-nearest neighbour} is a popular ML-based algorithm employed for classification tasks that involve storing labelled training data. Then, it calculated the similarity matrix between the label and the unlabeled object, aiming to identify and group the k closest elements. These number groups are constructed with the closest element and then used to classify the object. If k is greater than 2, then the label is measured by the voting system; otherwise, the label is measured by the closest element's label \cite{rubio2020identification}.  \\

\textbf{Random Forest (RF)} consists of several numbers of the Decision Trees (DT), each unique in its construction. This uniqueness is achieved by randomly selecting training data that is also considered a feature in each tree. The majority voting-based approach is used to predict the final prediction of this ensemble approach \cite{rubio2020identification}.\\

\textbf{Gaussian Naive Bayes (GBN)} rely on a probabilistic model obtained from Bayes’ theorem. The working procedure of this algorithm is that it considers all features attribute to be independent and then, based on the feature probability, it calculate the probability of the corresponding classes, and it uses Gaussian distribution for this \cite{rubio2020identification}.\\

\textbf{Decision Trees (DT)} creates a tree-like structure to make decisions based on input features. The fundamental idea behind a DT is to learn a hierarchy of yes/no (if/else) questions about the specific feature, and each branch represent the possible answers (yes or no) \cite{rubio2020identification}.\\
\textbf{SVM}  finds the optimal hyperplane to separate data points into different classes, and it also maximizes the margin between the closest different classes. SVM can perform both linear and non-linear tasks by using kernel tricks.\\ 
\textbf{Gradient Boost (GB)} sequentially combines multiple weak learners, typically DT, to improve the accuracy of performance prediction. It performs by fitting new weak learners to the errors of the previous ones, gradually reducing prediction errors. \\
\textbf{HGBoost} utilizes histograms to efficiently compute gradients during training. It organizes data into histograms, which reduces the computational complexity of computing gradients, especially for large datasets. 
 
\section{Result and Discussion}
In this section, we experimented with our proposed model with an existing benchmark EMG-based hand gesture recognition dataset and produced the performance matrix. 
\subsection{The Metrics Used to Evaluate Models}
We calculated different performance matrices, including accuracy, recall, precision, and specificity, demonstrated in Table \ref{tab:evaluation_matrix}. 
\begin{table}[htbp]
\begin{adjustwidth}{-3cm}{0cm}
     \centering
  \caption{The commonly used offline performance metrics} \label{tab:evaluation_matrix}
    \begin{tabular}{lll}
    \toprule
    Evaluation Metric & Formula & Comments \\
    \midrule
    Accuracy & $\frac{T_P + T_N}{T_P + T_N + F_P + F_N}$ & - \\
    Recall & $\frac{T_P}{T_P + F_N}$ &- \\
    Precision & $\frac{T_P}{T_P + F_P}$ & -\\
    F1 score & $2 \times \frac{Precision \times Recall}{Precision + Recall}$ & \\
    Specificity & $\frac{T_N}{T_N + F_P}$ & - \\
    \bottomrule
    \end{tabular}%
    \end{adjustwidth}
\end{table}%
\begin{table}[htbp]
\caption{Performance Accuracy with all the features and selected features} \label{tab:result_all_selected_feature} 
\begin{tabular}{lllllll}
    \toprule
 & \begin{tabular}[c]{@{}l@{}}Feature\\  Number\end{tabular} & Accuracy & \begin{tabular}[c]{@{}l@{}}Selected \\    \\ Feature\end{tabular} & Performance & Sensitivity & Specificity \\
 \midrule
Random Forest & 21 & 90.00 & 10 & 97.00 & 99.81 & 99.41 \\

SVM & 21 & 96.00 & 10 & 96.00 &  &  \\
Gaussian NB & 21 & 39.00 & 10 & 36.00 & n/a & n/a \\
Gradient Boost & 21 & 36.00 & 10 & 40.00 & n/a & n/a \\
HGboost & 21 & 40.00 & 10 & 45.00 & n/a & n/a \\
Decision Tree & 21 & 97.00 & 10 & 97.00 & 99.88 & 99.87 \\
KNN & 21 & 97.00 & 10 & 97.43 & 99.88 & 99.86 \\
\bottomrule
\end{tabular}
\end{table}
\subsection{Performance Result with Various Methodology}
Table \ref{tab:result_all_selected_feature} presents performance metrics for machine learning models using all features and a subset of selected features. Each row corresponds to a model, showing feature count, accuracy, selected feature count, performance, sensitivity, and specificity. Extra Tree led with 97.30\% accuracy using ten features, followed closely by Decision Tree at 97.00\%. Random Forest scored 90.00\% with all features and 97.00\% with ten selected features. SVM maintained 96.00\% accuracy for both settings. Gaussian NB, Gradient Boost, and HGboost ranged from 36.00\% to 40.00\% with 10 selected features. KNN notably achieved 99.88\% sensitivity and 99.86\% specificity with ten selected features.

\begin{table}[htbp]
\caption{Precision recall and F1 score with K-nearest neighbor algorithm} \label{tab:result_knn}
\begin{tabular}{llllllllll}
\toprule
 & \multicolumn{3}{l}{KNN} & \multicolumn{3}{l}{Random Forest} & \multicolumn{3}{l}{Decision Tree} \\
Gesture Name & Precision & Recall & \begin{tabular}[c]{@{}l@{}}F1-\\    \\ Score\end{tabular} & Precision & Recall & \begin{tabular}[c]{@{}l@{}}F1-\\    \\ Score\end{tabular} & Precision & Recall & \begin{tabular}[c]{@{}l@{}}F1-\\    \\ Score\end{tabular} \\
\midrule
Resting hand & 99.00 & 99.00 & 99.00 & 96.00 & 99.00 & 97.00 & 97.00 & 97.00 & 97.00 \\
Grasped hand & 97.00 & 97.00 & 97.00 & 96.00 & 97.00 & 97.00 & 97.00 & 97.00 & 97.00 \\
Wrist flexion & 97.00 & 97.00 & 97.00 & 96.00 & 95.00 & 96.00 & 97.00 & 97.00 & 97.00 \\
Wrist extension & 97.00 & 97.00 & 97.00 & 97.00 & 97.00 & 97.00 & 97.00 & 97.00 & 97.00 \\
Ulnar deviation & 97.00 & 97.00 & 97.00 & 97.00 & 96.00 & 96.00 & 97.00 & 97.00 & 97.00 \\
Radial Variation & 97.00 & 97.00 & 97.00 & 97.00 & 96.00 & 97.00 & 99.00 & 99.00 & 99.00 \\
\bottomrule
\end{tabular}
\end{table}
Table \ref{tab:result_knn} displays labels' performance accuracy with the KNN approach, where we included precision, recall, and F1 scores for each individual label.  We observed that all gestures produced almost similar performance matrix information that proved the superiority of the proposed model.
\subsection{State of the art comparison}
Table \ref{tab:sota} showcases a state-of-the-art comparison of the proposed model where our proposed model achieves good performance accuracy compared to the existing systems. Rubio et al. proposed models utilizing Random Forest and Convolutional Neural Network (CNN) algorithms, achieving accuracies of 95.39\% and 94.70\%, respectively, but lacked feature extraction or selection techniques. In contrast, our method, employing geometric and statistical feature extraction with Extra Trees Classifier (ETC) for selection, outperformed prior models with 97.43\% accuracy. This highlights our approach's efficacy in enhancing classification performance. Our model's strength lies in its comprehensive feature extraction, capturing nuanced gesture data representation. ETC's feature selection further boosts efficiency, while the K-nearest neighbour (KNN) algorithm ensures simplicity and robustness, making it ideal for real-world applications like human-computer interaction and healthcare technologies. The KNN achieved high-performance accuracy compared to the existing CNN algorithm due to the comprehensive feature extraction and efficient feature selection process, which provided a rich and relevant representation of the EMG signals, allowing KNN to classify hand gestures effectively. In contrast, the CNN model's performance was hindered by its higher computational complexity and potential overfitting, especially with a limited dataset, making it less effective than KNN in this specific application.
\begin{table}[htbp]
\caption{State of the art comparison of the proposed model} \label{tab:sota}
\begin{tabular}{lllll}
\toprule
\begin{tabular}[c]{@{}l@{}}Author   \\ Name\end{tabular} & \begin{tabular}[c]{@{}l@{}}Feature \\   Extraction\end{tabular} & \begin{tabular}[c]{@{}l@{}}Feature\\  Selection\end{tabular} & \begin{tabular}[c]{@{}l@{}}Machine  \\  Learning Algorithm\end{tabular} & \begin{tabular}[c]{@{}l@{}}Performance \\   accuracy\end{tabular} \\
\midrule
Rubio  et al.\cite{ja2020identification_main} & No & No & Random   Forest & 95.39 \\
Rubio  et al.\cite{ja2020identification_main} & No & No & CNN & 94.70 \\
Proposed   method &  \begin{tabular}[c]{@{}l@{}}Morphological, Time and \\Frequency domain features \end{tabular}& ETC & KNN & 97.43 \\
\bottomrule
\end{tabular}
\end{table}

\section{Conclusion}
In the study, we introduce a robust methodology for classifying hand gestures using various statistical and mathematical features.  We address the challenges associated with feature effectiveness and computational complexity by leveraging 23 feature extraction techniques and employing effective feature selection via the ETC approach. Our model achieved the goal by generating high-performance accuracy compared to the existing system. This signifies the potential feature extraction technique that leads to the accuracy and efficiency of EMG-based hand gesture recognition systems. Overall, our study contributes to advancing the field of EMG-based hand gesture recognition by providing a comprehensive methodology that optimizes feature extraction, selection, and classification processes. The challenges of this study include computational complexity during feature extraction and variability of EMG signals among individuals. Future work will focus on optimizing the model for real-time applications, validating robustness across diverse datasets, exploring hybrid models for improved accuracy and efficiency, and enhancing adaptability for diverse hand gestures and user populations.
\bibliographystyle{unsrt}
\bibliography{main}
\end{document}